\documentclass{article}

\PassOptionsToPackage{numbers, compress}{natbib}

\usepackage[final]{neurips_2024}

\usepackage[utf8]{inputenc} 
\usepackage[T1]{fontenc}    
\usepackage[hidelinks]{hyperref}       
\usepackage{url}            
\usepackage{booktabs}       
\usepackage{amsfonts}       
\usepackage{nicefrac}       
\usepackage{microtype}      
\usepackage{xcolor}         
\usepackage{multirow}
\usepackage{graphicx}
\usepackage{makecell}
\usepackage{subcaption}
\usepackage{enumitem}
\usepackage{xspace}

\newcolumntype{L}[1]{>{\raggedright\let\newline\\\arraybackslash\hspace{0pt}}m{#1}}
\newcolumntype{C}[1]{>{\centering\let\newline\\\arraybackslash\hspace{0pt}}m{#1}}
\newcolumntype{R}[1]{>{\raggedleft\let\newline\\\arraybackslash\hspace{0pt}}m{#1}}
\makeatletter 
\newcommand{\thickhline}{%
    \noalign {\ifnum 0=`}\fi \hrule height 1pt
    \futurelet \reserved@a \@xhline
}
\makeatother

\usepackage{listings}
\usepackage{xcolor}
\usepackage{float}
\newfloat{lstfloat}{htbp}{lop}
\floatname{lstfloat}{Listing}
\definecolor{commentsColor}{rgb}{0.497495, 0.497587, 0.497464}
\definecolor{keywordsColor}{rgb}{0.000000, 0.000000, 0.635294}
\definecolor{stringColor}{rgb}{0.558215, 0.000000, 0.135316}
\lstdefinestyle{mystyle}{
    basicstyle=\footnotesize\ttfamily,
    captionpos=b,
    breaklines=true,
    breakindent=0.5em,
    tabsize=2,
    frame=b,
    showstringspaces=false,
    numberstyle=\tiny\color{commentsColor},
    rulecolor=\color{black},
    commentstyle=\color{commentsColor}\textit,
    stringstyle=\color{stringColor},
    keywordstyle=\color{keywordsColor},
    emphstyle=\color{keywordsColor},
    escapeinside={(*@}{@*)},
}
\lstdefinestyle{tinyinline}{
    basicstyle=\scriptsize\ttfamily,
    captionpos=b,
    breaklines=true,
    breakindent=0.5em,
    tabsize=2,
    frame=b,
    showstringspaces=false,
    numberstyle=\tiny\color{commentsColor},
    rulecolor=\color{black},
    commentstyle=\color{commentsColor}\textit,
    stringstyle=\color{stringColor},
    keywordstyle=\color{keywordsColor},
    emphstyle=\color{keywordsColor},
    escapeinside={(*@}{@*)},
}
\AtBeginDocument{\DeclareCaptionSubType{lstlisting}}

\newcommand\ttc{TTC\xspace}
\newcommand\ctt{CTT\xspace}

\title{Don't Transform the Code, Code the Transforms: Towards Precise Code Rewriting using LLMs}

\author{
\vspace{2.5mm}
  Chris Cummins \hspace{6mm} Volker Seeker \hspace{6mm} Jordi Armengol-Estapé \hspace{6mm} Aram H. Markosyan \\
  \vspace{3mm}
  \textbf{Gabriel Synnaeve \hspace{10mm} Hugh Leather} \\
  Meta \\
  \vspace{3mm}
  \texttt{cummins@meta.com} \\
}

\begin{document}

\maketitle

\begin{abstract}

Tools for rewriting, refactoring and optimizing code should be fast and correct.
Large language models (LLMs), by their nature, possess neither of these qualities.
Yet, there remains tremendous opportunity in using LLMs to improve code.

We explore the use of LLMs not to \emph{transform code}, but to \emph{code transforms}.
We propose a chain-of-thought approach to synthesizing code transformations from a small number of input/output code examples that incorporates execution and feedback.
Unlike the direct rewrite approach, LLM-generated transformations are easy to inspect, debug, and validate.
The logic of the rewrite is explicitly coded and easy to adapt.
The compute required to run code transformations is minute compared to that of LLM rewriting.

We test our approach on 16 Python code transformations and find that LLM-generated transforms are perfectly precise for 7 of them and less imprecise than direct LLM rewriting on the others.
We hope to encourage further research to improving the precision of LLM code rewriting.
\end{abstract}

\section{Introduction}
\label{sec:intro}

A code transformation $f(c) \rightarrow c'$ is a function that rewrites input code $c$ to produce $c'$.
A multitude of software tasks can be expressed as code transformations from compiler optimizations to legacy code refactoring.
Traditional rule-based code transformations are challenging to implement, and there is increasing interest in using LLMs to estimate them~\citep{llm-swe-survey,llm-swe-agents-survey,llm-codegen-survey,pie,llm-compiler,supersonic,slade}.
However, in contrast to rule-based transformations, the logic of LLMs is opaque, provides no correctness guarantees, and is hard to debug when incorrect.
Instead, we propose using the LLM as a \emph{generator} for code transformations, $g(C, C') \rightarrow f$, where the output is an implementation of a code transformation, $f()$, inferred from input/output code examples $(C, C')$ before and after a transformation has been applied.

\begin{figure}[h!]
\begin{center}
\centerline{\includegraphics[width=1\textwidth]{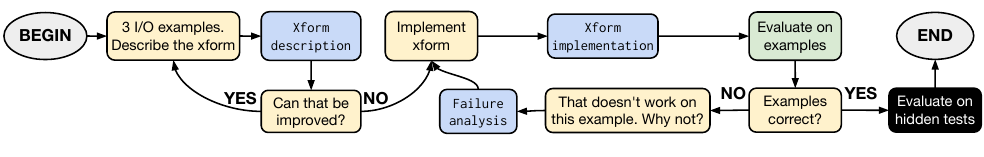}}
\vspace{-1em}
\caption{%
    Our chain-of-thought approach to synthesizing code transformations (xforms) using LLMs.
    Yellow boxes are prompts, blue are LLM outputs.
}
\label{fig:overview}
\end{center}
\end{figure}

\section{Synthesizing Code Transformations from Input/Output Examples}

We present a chain-of-thought~\citep{chain-of-thought} approach for the efficient synthesis of code transformations from input/output examples, shown in Figure~\ref{fig:overview}.
The approach requires only a small number of input/output examples to successfully generate a code transformation implementation.
Key to our approach is the plentiful use of loopback iterations to encourage the model to introspect on its own output, and to speculate as to the cause of failures before attempting to fix them.
The methodology is as follows:

\begin{enumerate}
\item We first present three input/output code examples and prompt the model to describe the underlying transform in precise natural language.
\item We allow the model to iterate on this description up to 10 times. We found that the initial description often under-explains the transformation, particularly the handling of edge cases.
\item Once the model assesses that the description is adequate, we provide this description, along with the original input/output code examples, and prompt the model to generate an implementation of the transform.
\item We take the model-generated transform implementation and execute it in a sandboxed environment against 10 input/output examples, including the ones it has seen. If the transform fails to produce the correct output, or if it crashes, we provide the counterexample or error message.
\item In case of failure we then perform an additional introspection step in which we ask the model to explain why the problem occurred.
\item We prompt the model with the previous incorrect code, the failing counterexample (both expected output and actual output), and the failure analysis generated in the previous step.
\item This process repeats until the transform works on all examples, up to a maximum of 50 iterations. We then evaluate the quality of the generated transform on unseen input/output examples.
\end{enumerate}

In this work we target Python and formulate code transformations as Abstract Syntax Tree (AST) rewrites using the format: \lstinline{def xform(code: ast.AST) -> ast.AST}.

\begin{table}[h]
\caption{
    Python code transformations, and the performance of two approaches: \emph{Transform the code}, and \emph{Code the transform}. We show F1 scores, with precision (how accurately the transformation is applied) and recall (how often transformation opportunities are identified) in parentheses.
    \label{tab:main}
}
\centering
\scriptsize
\lstset{style=tinyinline}  
\begin{tabular}{rlL{4.5cm}cc}
\toprule
Class & Transform & Description & \emph{Transform the code} & \emph{Code the transform} \\
\midrule
\multirow{3}{*}{Arithmetic}  & Add / subtract zero & Simplify $x + 0 \rightarrow x$ and $x - 0 \rightarrow x$. & 0.92 (0.85, 1.00) & \textbf{1.00} (1.00, 1.00)  \\
\cline{2-5}
 & Constant folding & Evaluate integer literal expressions in-place, e.g. $x = 10 + 15\rightarrow x = 25$. & 0.95 (0.91, 1.00) & \textbf{1.00} (1.00, 1.00)  \\
\cline{2-5}
 & Divide / multiply by one   & Simplify $x \div 1 \rightarrow x$ and $x \times 1 \rightarrow x$. & 0.93 (0.88, 1.00) & \textbf{1.00} (1.00, 1.00)  \\
\midrule
\multirow{3}{*}{Boolean}  & Collapse nested ifs & Recursively flatten nested \lstinline!if! conditionals to a compound conditional. & 0.81 (0.68, 1.00) & \textbf{1.00} (1.00, 1.00)  \\
\cline{2-5}
 & De Morgan's law & Rewrite \lstinline;!(a & b); $\rightarrow$ \lstinline;!a | !b;. & 0.82 (0.69, 1.00) & \textbf{1.00} (1.00, 1.00)  \\
\cline{2-5}
 & Reorder conditional & Flip the branches in \lstinline!if not/else! conditionals to \lstinline!if/else!. & 0.52 (0.35, 1.00) & \textbf{0.93} (0.86, 1.00)  \\
\midrule
\multirow{3}{*}{Liveness}  & Dead code elimination & Remove \texttt{if} conditionals if the branch condition statically evaluates to \texttt{False}. & 0.93 (0.88, 1.00) & \textbf{0.99} (0.99, 0.99)  \\
\cline{2-5}
 & Redundant fn.\ elimination & Remove function definitions, and their calls, if the function contains no instructions. & 0.93 (0.87, 1.00) & \textbf{1.00} (1.00, 1.00)  \\
\cline{2-5}
 & Unused var.\ elimination & Remove declared but unused variables. & 0.87 (0.77, 1.00) & \textbf{0.98} (0.96, 1.00)  \\
\midrule
\multirow{3}{*}{Loops}  & List comprehension & Rewrite \lstinline!for! loop as list comprehension. & 0.60 (0.43, 1.00) & \textbf{0.90} (0.86, 0.95)  \\
\cline{2-5}
 & List comp.\ w.\ condition & As above but the loop body has a conditional. & 0.62 (0.45, 1.00) & \textbf{0.82} (0.73, 0.93)  \\
\cline{2-5}
 & Loop dupe & Duplicate loops (not semantics preserving). & 0.50 (0.34, 1.00) & \textbf{0.99} (1.00, 0.99)  \\
\cline{2-5}
 & Loop unroll & Fully unroll loops with statically known \lstinline!range()! iteration bounds. & 0.82 (0.70, 1.00) & \textbf{0.98} (0.99, 0.98)  \\
\midrule
\multirow{3}{*}{Optimization}  & Dot product to torch & Replace \lstinline!for! loop that computes vector dot product with torch API. & 0.66 (0.49, 1.00) & \textbf{0.94} (0.95, 0.93)  \\
\cline{2-5}
 & Pointwise add to torch & Replace \lstinline!for! loop that computes pointwise add with torch API. & 0.57 (0.40, 1.00) & \textbf{0.97} (0.94, 1.00)  \\
\cline{2-5}
 & Torch zero grad & Replace \lstinline!m.zero_grad()! with a loop over model parameters, assigning to \lstinline!None!. & 0.87 (0.77, 1.00) & \textbf{1.00} (1.00, 1.00)  \\
\midrule
\multicolumn{3}{l}{\textbf{Overall}} & 0.75 (0.60, 1.00) & \textbf{0.97} (0.95, 0.99) \\
\bottomrule
\end{tabular}
\lstset{style=mystyle}  
\end{table}

\begin{figure}
\begin{center}
\centerline{\includegraphics[width=\textwidth]{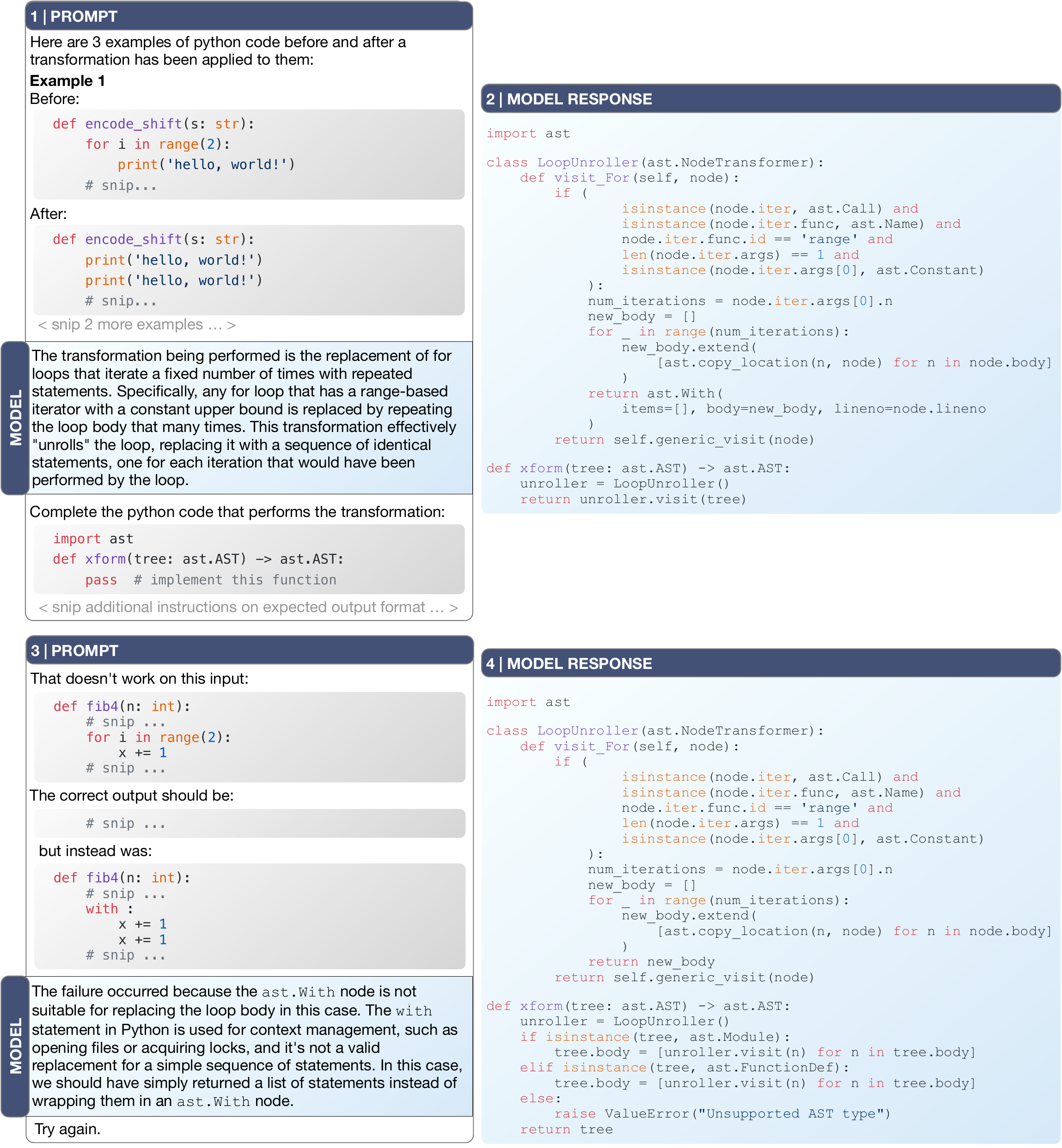}}
\caption{%
    An example LLM dialog, showing how the results of previous queries are used to generate natural language descriptions and failure analyses that are integrated into the chain-of-thought.
}
\label{fig:dialog}
\end{center}
\end{figure}

\begin{figure}
\centering
\centerline{\includegraphics[width=1\textwidth]{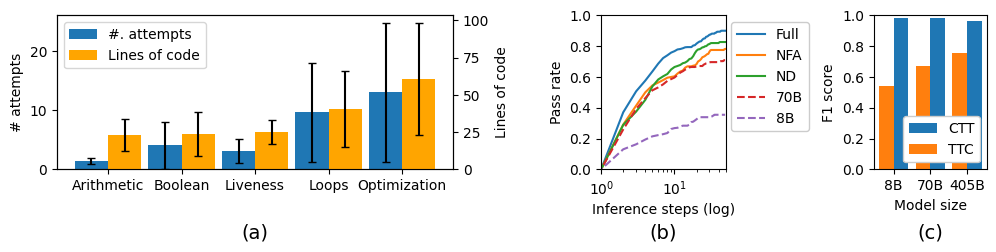}}
\vspace{-.5em}
\caption{%
    (a) the average number of attempts needed to synthesize transforms of each class correlates with the size of the transforms; (b) more attempts are required if chain-of-thought steps are removed or smaller models used; (c) synthesized transforms (\ctt) of all model sizes work similarly well, but code rewriting ability (\ttc) scales with model size.
}
\label{fig:results}
\end{figure}

\newpage
\section{Experiments}

We evaluate the performance of our \emph{Code the transforms} approach (\ctt) and compare it against a \emph{Transform the code} (\ttc) approach in which the LLM is used to directly rewrite the code.
For \ttc we provide the model 10 examples and prompt it to apply the same transformation to unseen codes in turn.
For \ctt we use the examples to synthesize a transform in the manner described previously and then test the synthesized transform against unseen codes.
We use the 405B parameter Llama 3.1 model~\cite{llama3} and sample it with temperature 0.25. We repeat the experiment 10 times.

\paragraph{Benchmarks.} We assemble a dataset of 480 input/output Python programs covering 16 code transformations, summarized in Table~\ref{tab:main}.
We aim to cover a range of uses and complexities from simple semantics-preserving rewrites (e.g.\ \emph{constant folding}) to more substantial code optimizations (e.g.\ \emph{dot product to torch}).
We generated the example programs by adapting HumanEval solutions~\citep{humaneval}.

Our benchmark comprises 30 input/output Python code pairs for each of the transformations: 10 public examples available to the LLM, 10 hidden examples, and a further 10 examples of code where the transformation does not apply. The average length of each program is 11 lines of code.

\paragraph{Metrics.} Code transformation requires that two tasks be completed successfully: \emph{identifying} regions of code eligible for transform, and \emph{executing} the transformation on identified code regions.
To assess these properties we use the binary metrics of Precision and Recall. For a particular \texttt{(input,expected\_out,actual\_out)} tuple, the transform is \emph{precise} if \lstinline!actual_out == expected_out!, and successfully \emph{recalled} if \lstinline;actual_output != input && expected_out != input;. F1 is the harmonic mean of precision and recall. Scores are calculated over 10 runs.

\paragraph{Results.} Table~\ref{tab:main} compares the performance of both approaches. CTT has higher precision than \ttc across every problem (overall 0.95 vs 0.60).
Although overall scores are high, \ctt still occasionally fails.
Figure~\ref{fig:dialog} shows an example failure and the model self-correcting.
Qualitatively, we found that while \ctt does a good job at comprehending the problem, it often struggles to turn that into working code.
For example, the simple \emph{arithmetic} transforms required only 1.5 attempts on average to synthesize a transform, whereas \emph{optimization} transforms, which require more substantial code changes, require 11.8 (Figure~\ref{fig:results}a).
Typically we found errors made by \ctt to be easy to debug.
Synthesized transforms average 34.4 lines of code (Figure~\ref{fig:results}a), compared to \ttc which requires reviewing every output to check for errors. \ctt performs slightly worse than \ttc on recall (0.99 vs 1.00), as it would occasionally miss opportunities to apply a transformation that can be applied.

\paragraph{Ablations.} Figure~\ref{fig:results}b compares the rate at which code transformations are synthesized when using our full chain-of-thought approach (Full), when the failure analysis step is removed (NFA), and when the natural language description step is removed (ND). Removing these steps from the chain-of-thought reduces the rate of transform synthesis efficiency.

We also repeated the full chain-of-though experiments using the smaller 70B and 8B parameter Llama 3.1 variants. Interestingly, we discovered that while the smaller models require many more inferences to synthesize transforms (Figure~\ref{fig:results}b), the transforms synthesized by all models perform equally well (Figure~\ref{fig:results}c), suggesting a compute/inference budget tradeoff.
For \ttc we see the expected result that smaller models are worse at directly rewriting code.

\section{Discussion}

We cannot afford the burden of reviewing and testing every piece of code an LLM touches. We propose an alternative formulation that reduces reviewing and testing costs by having the LLM instead generate code transformations. We show that this is more precise than using the LLM directly, but there is still a way to go. For example, we see in Figure~\ref{fig:dialog} that although the model-generated code provides the expected transformation, there are obvious improvements that a human developer would make. We suspect that
reinforcement learning and bootstrapped fine-tuning~\citep{star} could improve performance.

\clearpage
\bibliography{references}

\begin{thebibliography}{11}
\providecommand{\natexlab}[1]{#1}
\providecommand{\url}[1]{\texttt{#1}}
\expandafter\ifx\csname urlstyle\endcsname\relax
  \providecommand{\doi}[1]{doi: #1}\else
  \providecommand{\doi}{doi: \begingroup \urlstyle{rm}\Url}\fi

\bibitem[Hou et~al.(2023)Hou, Zhao, Liu, Yang, Wang, Li, Luo, Lo, Grundy, and
  Wang]{llm-swe-survey}
Xinyi Hou, Yanjie Zhao, Yue Liu, Zhou Yang, Kailong Wang, Li~Li, Xiapu Luo,
  David Lo, John Grundy, and Haoyu Wang.
\newblock {Large Language Models for Software Engineering: A Systematic
  Literature Review}.
\newblock \emph{arXiv:2308.10620}, 2023.

\bibitem[Liu et~al.(2024)Liu, Wang, Chen, Peng, Chen, Zhang, and
  Lou]{llm-swe-agents-survey}
Junwei Liu, Kaixin Wang, Yixuan Chen, Xin Peng, Zhenpeng Chen, Lingming Zhang,
  and Yiling Lou.
\newblock {Large Language Model-Based Agents for Software Engineering: A
  Survey}.
\newblock \emph{arXiv:2409.02977}, 2024.

\bibitem[Jiang et~al.(2024)Jiang, Wang, Shen, Kim, and Kim]{llm-codegen-survey}
Juyong Jiang, Fan Wang, Jiasi Shen, Sungju Kim, and Sunghun Kim.
\newblock {A Survey on Large Language Models for Code Generation}.
\newblock \emph{arXiv:2406.00515}, 2024.

\bibitem[Madaan et~al.(2023)Madaan, Shypula, Alon, Hashemi, Ranganathan, Yang,
  Neubig, and Yazdanbakhsh]{pie}
Aman Madaan, Alexander Shypula, Uri Alon, Milad Hashemi, Parthasarathy
  Ranganathan, Yiming Yang, Graham Neubig, and Amir Yazdanbakhsh.
\newblock {Learning Performance-Improving Code Edits}.
\newblock \emph{arXiv:2302.07867}, 2023.

\bibitem[Cummins et~al.(2024)Cummins, Seeker, Grubisic, Roziere, Gehring,
  Synnaeve, and Leather]{llm-compiler}
Chris Cummins, Volker Seeker, Dejan Grubisic, Baptiste Roziere, Jonas Gehring,
  Gabriel Synnaeve, and Hugh Leather.
\newblock {Meta Large Language Model Compiler: Foundation Models of Compiler
  Optimization}.
\newblock \emph{arXiv:2407.02524}, 2024.

\bibitem[Chen et~al.(2024)Chen, Fang, and Monperrus]{supersonic}
Zimin Chen, Sen Fang, and Martin Monperrus.
\newblock {Supersonic: Learning to generate source code optimizations in
  C/C++}.
\newblock \emph{TSE}, 2024.

\bibitem[Armengol-Estap{\'e} et~al.(2024)Armengol-Estap{\'e}, Woodruff,
  Cummins, and O'Boyle]{slade}
Jordi Armengol-Estap{\'e}, Jackson Woodruff, Chris Cummins, and Michael~FP
  O'Boyle.
\newblock {SLaDe: A Portable Small Language Model Decompiler for Optimized
  Assembler}.
\newblock In \emph{CGO}, 2024.

\bibitem[Wei et~al.(2022)Wei, Wang, Schuurmans, Bosma, Xia, Chi, Le, Zhou,
  et~al.]{chain-of-thought}
Jason Wei, Xuezhi Wang, Dale Schuurmans, Maarten Bosma, Fei Xia, Ed~Chi, Quoc~V
  Le, Denny Zhou, et~al.
\newblock {Chain-of-Thought Prompting Elicits Reasoning in Large Language
  Models}.
\newblock In \emph{NeurIPS}, 2022.

\bibitem[Dubey et~al.(2024)Dubey, Jauhri, Pandey, Kadian, Al-Dahle, Letman,
  Mathur, Schelten, Yang, Fan, et~al.]{llama3}
Abhimanyu Dubey, Abhinav Jauhri, Abhinav Pandey, Abhishek Kadian, Ahmad
  Al-Dahle, Aiesha Letman, Akhil Mathur, Alan Schelten, Amy Yang, Angela Fan,
  et~al.
\newblock {The Llama 3 Herd of Models}.
\newblock \emph{arXiv:2407.21783}, 2024.

\bibitem[Chen et~al.(2021)Chen, Tworek, Jun, Yuan, de~Oliveira~Pinto, Kaplan,
  Edwards, Burda, Joseph, Brockman, Ray, Puri, Krueger, Petrov, Khlaaf, Sastry,
  Mishkin, Chan, Gray, Ryder, Pavlov, Power, Kaiser, Bavarian, Winter, Tillet,
  Such, Cummings, Plappert, Chantzis, Barnes, Herbert-Voss, Guss, Nichol,
  Paino, Tezak, Tang, Babuschkin, Balaji, Jain, Saunders, Hesse, Carr, Leike,
  Achiam, Misra, Morikawa, Radford, Knight, Brundage, Murati, Mayer, Welinder,
  McGrew, Amodei, McCandlish, Sutskever, and Zaremba]{humaneval}
Mark Chen, Jerry Tworek, Heewoo Jun, Qiming Yuan, Henrique~Ponde
  de~Oliveira~Pinto, Jared Kaplan, Harri Edwards, Yuri Burda, Nicholas Joseph,
  Greg Brockman, Alex Ray, Raul Puri, Gretchen Krueger, Michael Petrov, Heidy
  Khlaaf, Girish Sastry, Pamela Mishkin, Brooke Chan, Scott Gray, Nick Ryder,
  Mikhail Pavlov, Alethea Power, Lukasz Kaiser, Mohammad Bavarian, Clemens
  Winter, Philippe Tillet, Felipe~Petroski Such, Dave Cummings, Matthias
  Plappert, Fotios Chantzis, Elizabeth Barnes, Ariel Herbert-Voss,
  William~Hebgen Guss, Alex Nichol, Alex Paino, Nikolas Tezak, Jie Tang, Igor
  Babuschkin, Suchir Balaji, Shantanu Jain, William Saunders, Christopher
  Hesse, Andrew~N. Carr, Jan Leike, Josh Achiam, Vedant Misra, Evan Morikawa,
  Alec Radford, Matthew Knight, Miles Brundage, Mira Murati, Katie Mayer, Peter
  Welinder, Bob McGrew, Dario Amodei, Sam McCandlish, Ilya Sutskever, and
  Wojciech Zaremba.
\newblock {Evaluating Large Language Models Trained on Code}.
\newblock \emph{arXiv:2107.03374}, 2021.

\bibitem[Zelikman et~al.(2022)Zelikman, Wu, Mu, and Goodman]{star}
Eric Zelikman, Yuhuai Wu, Jesse Mu, and Noah Goodman.
\newblock {STaR: Bootstrapping Reasoning with Reasoning}.
\newblock \emph{NeurIPS}, 2022.

\end{thebibliography}

\end{document}